\documentclass[]{bytedance_seed}

\usepackage[toc,page,header]{appendix}

\usepackage{minitoc}

\title{BootSeer: Analyzing and Mitigating Initialization Bottlenecks in Large-Scale LLM Training}

\author[1,2,*, \dagger]{Rui Li}
\author[1]{Xiaoyun Zhi}
\author[1]{Jinxin Chi}
\author[1]{Menghan Yu}
\author[3]{Lixin Huang}
\author[3]{Jia Zhu}
\author[3]{Weilun Zhang}
\author[3]{Xing Ma}
\author[3]{Wenjia Liu}
\author[3]{Zhicheng Zhu}
\author[3]{Daowen Luo}
\author[1, \dagger]{Zuquan Song}
\author[3]{Xin Yin}
\author[3]{Chao Xiang}
\author[1]{Shuguang Wang}
\author[1]{Wencong Xiao}
\author[2]{Gene Cooperman}

\affiliation[1]{ByteDance Seed}
\affiliation[2]{Northeastern University}
\affiliation[3]{ByteDance}

\contribution[*]{Work done at ByteDance Seed}
\contribution[\dagger]{Corresponding authors}

\abstract{
Large Language Models (LLMs) have become a cornerstone of modern AI, driving breakthroughs in natural language processing and expanding into multimodal jobs involving images, audio, and video.
As with most computational software, it is important to distinguish between ordinary runtime performance and startup overhead.
Prior research has focused on runtime performance: improving training efficiency and stability.
This work focuses instead on the increasingly critical issue of startup overhead in training: the delay before training jobs begin execution.
Startup overhead is particularly important in large, industrial-scale LLMs, where failures occur more frequently and multiple teams operate in iterative update-debug cycles.
In one of our production training clusters, startup overhead alone wastes over 4{,}700 GPU-hours in a single day.

In this work, we present the first in-depth characterization of LLM training startup overhead based on real production data. 
We analyze the components of startup cost, quantify its direct impact, and examine how it scales with job size. 
These insights motivate the design of Bootseer, a system-level optimization framework that addresses three primary startup bottlenecks: (a) container image loading, (b) runtime dependency installation, and (c) model checkpoint resumption. 
To mitigate these bottlenecks, Bootseer introduces three techniques: (a) hot block record-and-prefetch, (b) dependency snapshotting, and (c) striped HDFS-FUSE. Bootseer has been deployed in a production environment and evaluated on real LLM training workloads, demonstrating a 50\% reduction in startup overhead.
}

\date{\today}
\correspondence{Author1 at \email{li.rui8@northeastern.edu}, Author12 at \email{zuquan.song@bytedance.com}}

\begin{document}
\maketitle
\section{Introduction}
\label{sec:Intro}
Large Language Models (LLMs) have emerged as one of the most prominent developments in both industry and academia in recent years~\cite{bommasani2021opportunities}. 
Their ability to understand and generate human language has led to significant breakthroughs in Natural Language Processing (NLP)~\cite{brown2020language}. 
Today, LLMs are extending beyond text to handle multimodal jobs involving images, audio, and video, enabling applications in areas such as autonomous driving and robotics. 
The vast potential of LLMs has drawn major tech companies into a competitive race to develop their own models—examples include ChatGPT~\cite{achiam2023gpt}, Gemini~\cite{team2023gemini}, Grok~\cite{xai2023grok}, DeepSeek~\cite{liu2024deepseek}, and others. 
These companies are building massive GPU clusters for both training and inference, accelerating the race toward the next major leap in AI capabilities.

The development of LLMs highlights the critical role of scaling laws~\cite{kaplan2020scaling} in driving performance improvements. 
For instance, GPT-3~\cite{floridi2020gpt} contains 175 billion parameters, while PaLM~\cite{chowdhery2023palm} scales up to 540 billion parameters,
and there is speculation that GPT-4 has more than a trillion parameters. 
Compared to earlier deep learning models such as BERT~\cite{devlin2019bert}, recent LLMs have not significantly changed in architectural design; instead, their performance gains largely stem from increased model size, larger training datasets, and vastly greater computational resources. 
Training has scaled from using tens of GPUs to thousands or more~\cite{jiang2024megascale}. 
At least in the near future, scaling laws are expected to remain a central factor in the advancement of LLM capabilities.

The scaling laws that drive LLM performance also introduce significant challenges for system design, inspiring a new wave of machine learning systems research. 
Previous research has primarily focused on improving both the efficiency and stability of large-scale LLM training. 
To enhance training efficiency, researchers have optimized communication overhead, operator performance, data preprocessing pipelines, and other system-level components~\cite{narayanan2021efficient, jiang2024megascale, wan2025bytecheckpoint}. 
In terms of stability, existing work addresses issues such as fault tolerance and performance degradation caused by hardware failures or straggling nodes during distributed training~\cite{wu2024falcon, dong2024boosting, cui2025xputimer,xiong2024superbench}. 
Given the high cost of GPUs, improving efficiency and stability reduces training expenses and accelerates LLM development.

However, training startup overhead, once considered negligible, has become a significant source of resource inefficiency in large-scale LLM training.
Although a startup delay of several minutes seems trivial if one naively assumes a single training job that extends over several weeks, real-world training workflows are far more dynamic. 
Training an LLM is not simply a matter of launching a job on a massive GPU cluster and letting it run uninterrupted for weeks. 

In practice, large organizations often run many training jobs simultaneously, managed by sizable algorithm teams.
These jobs frequently stop and restart due to debugging, system failures, node slowdowns, or iterative algorithm updates.
For example, a recent training job on our platform running on 4{,}864 NVIDIA H800 GPUs experienced 26 full startups (10 due to debugging) and 6 hot updates within a 21-hour window.

Although each startup is transient, their cumulative impact is substantial.
Slow startup phases significantly degrade development efficiency by lengthening the debug–modify–resubmit cycle.
Instead of iterating rapidly on model or system changes, engineers must repeatedly wait for long initialization phases before observing any training progress.
This delay slows experimentation, reduces effective developer throughput, and exacerbates coordination overhead across large teams.
Moreover, prolonged and repeated initialization increases exposure to transient system issues, further disrupting the development process.

When discussing startup overhead, it is important to note that extensive research has been conducted on startup delays in smaller-scale systems in other domains, particularly in the serverless computing domain~\cite{du2020catalyzer,brooker2023demand,wei2023no,cadden2020seuss}. 
Researchers and engineers have proposed various optimizations for different startup phases, such as image loading~\cite{brooker2023demand} and dependency loading~\cite{oakes2018sock}. 
Techniques like lazy loading, caching~\cite{brooker2023demand}, page pre-fetching~\cite{ustiugov2021benchmarking}, and checkpointing~\cite{du2020catalyzer} have shown strong results in this area. 
Some of these strategies, such as image lazy loading, caching, may also be applicable to LLM training. 
However, others, like transparent checkpointing of the entire application state (\hbox{e.g.} CRIU~\cite{criu2019}), remain impractical due to the complexity and scale of LLM workloads.

However, in the domain of industrial-scale LLMs, the problem of startup overhead remains poorly characterized and introduces unique challenges. 
The startup process typically involves resource scheduling, image loading, environment initialization, launching supporting services, establishing TCP and RDMA connections, and resuming from checkpoints. 
To alleviate startup overhead, it is essential to conduct a detailed analysis of the startup process in production environments and identify the primary bottlenecks.

In this paper, we present the first characterization of LLM training startup overhead based on production data, covering more than 28,000 training jobs that, in total, were assigned more than 70,000 GPUs across one week, accounting for reused and overlapping usage.
We aim to answer the following key questions: 
(i) How much GPU time is wasted due to startups during training, and how does it impact the training efficiency? 
(ii) How does each component of the startup process contribute to the total overhead? 
(iii) How does startup overhead scale with training job size? 
The answers to these questions provide insights into improving startup overhead and they highlight the importance of startup in enhancing the overall efficiency of LLM training.

We then propose \textbf{BootSeer}, a systematic optimization framework designed to reduce LLM training startup overhead. 
BootSeer first implements a profiler to collect and analyze the startup stages during training, so that we can collect and analyze the startup overhead.
BootSeer targets three key bottlenecks: (1)~concurrent loading of large container images; (2)~complex and redundant dependency installation; and (3)~resuming from a model checkpoint. 
While similar techniques exist in other domains, BootSeer adapts these techniques to a more specialized use during startups in LLM training: reducing startup bottlenecks for two common workloads: long jobs that restart often due to failures or debugging; and many short, similar jobs used for feature testing.

To address these challenges, BootSeer employs a combination of caching, prefetching, and peer-to-peer sharing strategies.
For image loading, it adopts block-level lazy loading similar to techniques used by AWS Lambda~\cite{brooker2023demand} and Nydus~\cite{nydus2025} as baseline.
BootSeer further introduces a record-and-prefetch mechanism that identifies hot blocks, prioritizes their loading to enable early startup, and streams remaining blocks in the background.
To eliminate redundant dependency setup, it creates job-level dependency snapshots, avoiding reinstallation during job restarts or node replacements.
Finally, BootSeer accelerates concurrent checkpoint resumption by implementing a striped HDFS-FUSE (Filesystem in Userspace) design, which achieves parallel read/write throughput and overlaps local I/O with HDFS download and upload operations.

BootSeer has been implemented and deployed in our production systems, and evaluated using a representative MOE model training workload. 
The experimental workload runs on up to 128 NVIDIA H800 GPUs. 
Experimental results show that BootSeer reduces startup overhead by 50\%, demonstrating its effectiveness in improving system responsiveness and resource efficiency during job initialization. 
Additionally, BootSeer effectively eliminates straggler effects as job scale increases.

We summarize the contributions as follows.
\begin{enumerate}
\item
We present the first in-depth study of startup overhead in LLM training, analyzing its components, bottlenecks, and case studies using real production data. We also explain why this overhead critically impacts training efficiency and system stability.
\item 
We propose \textbf{BootSeer}, a production-ready framework that mitigates startup overhead by leveraging characteristics of LLM training workflows and the underlying hardware environment.
\item 
We evaluate BootSeer across multiple LLM training jobs, demonstrating the effectiveness of each optimization technique in reducing startup time and improving overall system performance.
\end{enumerate}

The rest of this paper is organized as follows: 
Section~\ref{sec:background} provides background on the industrial-scale training system for LLMs, and with special emphasis on the startup process. 
Section~\ref{sec:motivation} presents a study and analysis of the startup overhead for the LLM training models. 
Section~\ref{sec:design} discusses the design choices and implementation details of BootSeer. 
Section~\ref{sec:eval} evaluates BootSeer on these full-scale LLM training jobs. 
We review related work in Section~\ref{sec:related_work} and conclude the paper in Section~\ref{sec:conclusion}.
\section{Background}
\label{sec:background}
This section provides background on large language model (LLM) training systems and outlines the startup process of LLM training jobs.

Section~\ref{sec:training} discusses the process of training and releasing a new LLM in a large organization.  
Section~\ref{sec:back_startup} then goes on to discuss the internals of the startup process.  This includes: a Scheduler Phase; allocation of resources; loading an image; setting up the environment; and initializing the execution.
Finally, Section~\ref{sec:back_system} points out where are the largest opportunities (``the low-hanging fruit'') for optimizing the startup phase in LLMs.

\subsection{LLM Training in Production}
\label{sec:training}
To motivate the following sections, we first review how a large industrial organization produces a new LLM. 
This contrasts with the academic setting, where researchers typically explore novel training algorithms using much smaller models and datasets. 
Our focus is on the significant startup overhead observed in large-scale, industrial LLM training --- an issue not representative of smaller models or datasets.

A simplistic conception of building an LLM model posits that if there is no failure of computer nodes, then the computation has just one startup phase at the beginning.
The remaining computation is simply to continue training until the loss for the LLM model is sufficiently small.
In this simplistic conception, the only reason to execute an additional startup phase is if there is a node failure, and the computation must revert to an earlier checkpoint.

At industrial scale, the reality is quite different. Training jobs may be terminated for several reasons, including:
\begin{itemize}
    \item The model experienced a software or hardware failure.
    \item The training was previously progressing well, but the loss suddenly increased for unknown reasons. In such cases, the job is terminated.  The model is then reverted to a checkpoint where the loss was still decreasing, and restarted multiple times for debugging.
    \item A team is experimenting with different post-training strategies. Multiple developer teams may apply various strategies to a base model. If a new approach fails to quickly produce improvements, the job is terminated and the next idea is tried.
    \item A higher-priority job has claimed the node resources, requiring the lower-priority job to be aborted.
\end{itemize}

\begin{figure}[!ht]
  \centering
  \includegraphics[width=0.45\columnwidth]{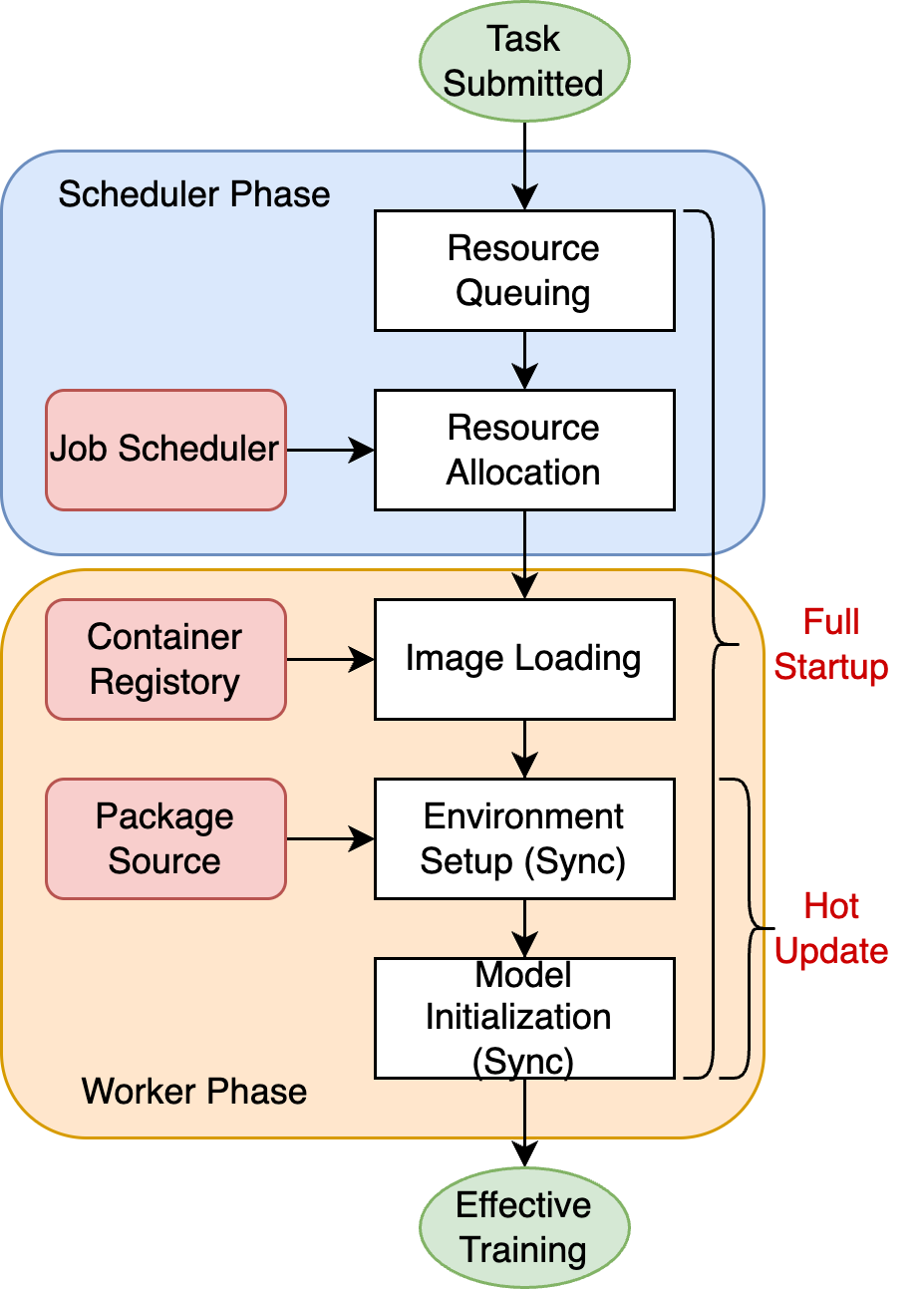}
  \caption{Startup process of an LLM training job. Steps marked with (Sync) indicate that all worker nodes must synchronize at that stage.}
  \label{fig:llm_startup}
\end{figure}

\subsection{Types of Startups}
\label{sec:back_startup}
Startups happen multiple times and for multiple reasons during an LLM training life-cycle, especially for those jobs using large-scale GPU cards.
A \emph{Full Startup} happens when a user: (i)~first submits a training job; (ii)~replaces a machine due to runtime failure/slowdowns; (iii)~debugs the software; or (iv)~changes a runtime configuration.
Meanwhile, a \emph{Hot Update} also causes overhead due to a \emph{Partial Startup} (setting up an environment again and rerunning the model setup).

Figure~\ref{fig:llm_startup} illustrates the simplified startup process of an LLM training job.  
In the \textit{Scheduler Phase}, the cluster operation appears similar to that of other distributed workloads.  
When a user submits a job, the scheduler provisions the required resources.  
Each worker node then ``pulls'' the necessary container images, starts the container, and executes the setup or entrypoint script.  
Next, each machine initializes the model and begins the training process.

\textbf{Resource Queuing.}
Jobs remain in a queue until their resource requirements are met and no higher-priority jobs are pending.  

\textbf{Resource Allocation.}  
The job scheduler (e.g., Kubernetes~\cite{carrion2022kubernetes,senjab2023survey}) is responsible for allocating resources for training jobs.  
Once sufficient resources are available --- typically a set of physical GPU machines --- the scheduler allocates them to the job and initiates the process of ``pulling'' the required container images.

The startup process then enters the \textit{Worker Phase}:

\noindent
\textbf{Image Loading.} 
In this stage, all worker nodes (GPU machines) begin pulling images concurrently from the container registry, and then start the containers.
These images are typically much larger than those used in smaller jobs --- ranging from 25 to 40~GB in our system.  
When the number of worker nodes is large, concurrent image loading places significant strain on both the network bandwidth and the container registry service.  
If any sidecar containers are required (e.g., HDFS-FUSE), their images are also pulled during this stage.

All machines must then wait for the slowest node to complete the image pull, in order to synchronize.
Sometimes, a worker node may already have the training image cached locally, eliminating the need to pull it from the remote registry.
Nevertheless, a cache hit on a single machine does not necessarily improve the overall efficiency at this stage.
All experiments in this paper adopt the lazy-loading strategy instead of using standardized OCI images.

\textbf{Environment Setup.}  
In this stage, each worker node prepares its environment before running the actual training program (typically a Python program).  
First, it installs any dependencies that are not included in the container image.  
This is often necessary due to heterogeneous environments and the flexibility required by engineers.  
For instance, the version of certain dependencies may vary depending on the machine type, GPU type, operating system, or even geographic region.  
And some dependencies change often, because engineers frequently update them, without rebuilding the container image.

Second, this stage also performs health checks and launches necessary daemon processes --- for example, those related to monitoring and performance analysis.  
Some of these daemons require synchronization across machines at this stage.  
Once again, if a single machine is slower than the others, all machines must wait, introducing potential startup delays.

\textbf{Model Initialization.}  
In this stage, the user program begins execution, but effective model training has not yet started.  
In this stage, several steps depend on the model architecture (e.g., Dense, MoE, RL, etc.).  
This involves setting up the model according to the parallelization strategy, launching ranks (i.e., processes running on each GPU), loading model checkpoints, initializing RDMA connections, among other steps.  

\textbf{Startup Cost Breakdown in Practice.}
To illustrate the cost of the above startup phases in real production settings, we report measurements from two recent industrial LLM training jobs.
All timings are measured at the job level.

The first job trains a pure language MoE base model on 1{,}440 machines (11{,}520 NVIDIA GPUs).
The initial worker process launch takes 21 seconds.
Model initialization (including imports, model graph construction, parallelism setup, and NCCL initialization) takes 41 seconds, within which NCCL initialization alone accounts for 6.7 seconds.
Resuming from a checkpoint takes 3 minutes and 13 seconds, and data loader setup takes 18.5 seconds.

The second job trains a multimodal model on 384 machines, with a substantially larger checkpoint that includes vision model parameters.
In this case, the initial worker process launch takes 10 seconds, model initialization takes 22 seconds, checkpoint resume takes 10 minutes and 12 seconds, and data loader setup takes 5 minutes and 17 seconds.
These results show that startup overheads can vary significantly across models and configurations, and that checkpoint and data loading phases can dominate the end-to-end startup cost.

\begin{figure*}[h]
  \centering
  \begin{subfigure}[t]{0.45\linewidth}
    \includegraphics[width=\linewidth]{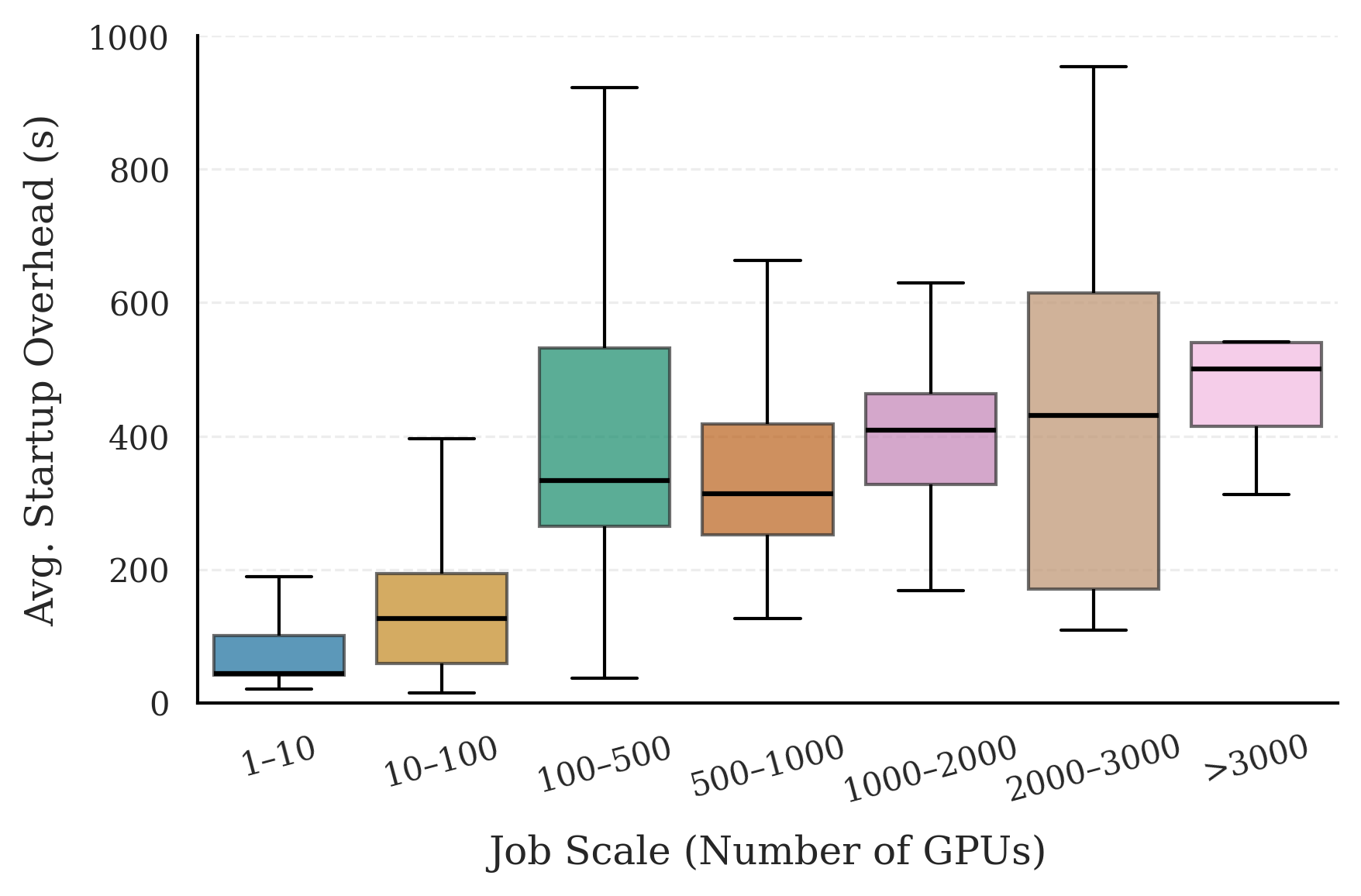}
    \caption{Job-level startup overhead.}
    \label{fig:trial_avg_startup_overhead}
  \end{subfigure}
  \hfill
  \begin{subfigure}[t]{0.45\linewidth}
    \includegraphics[width=\linewidth]{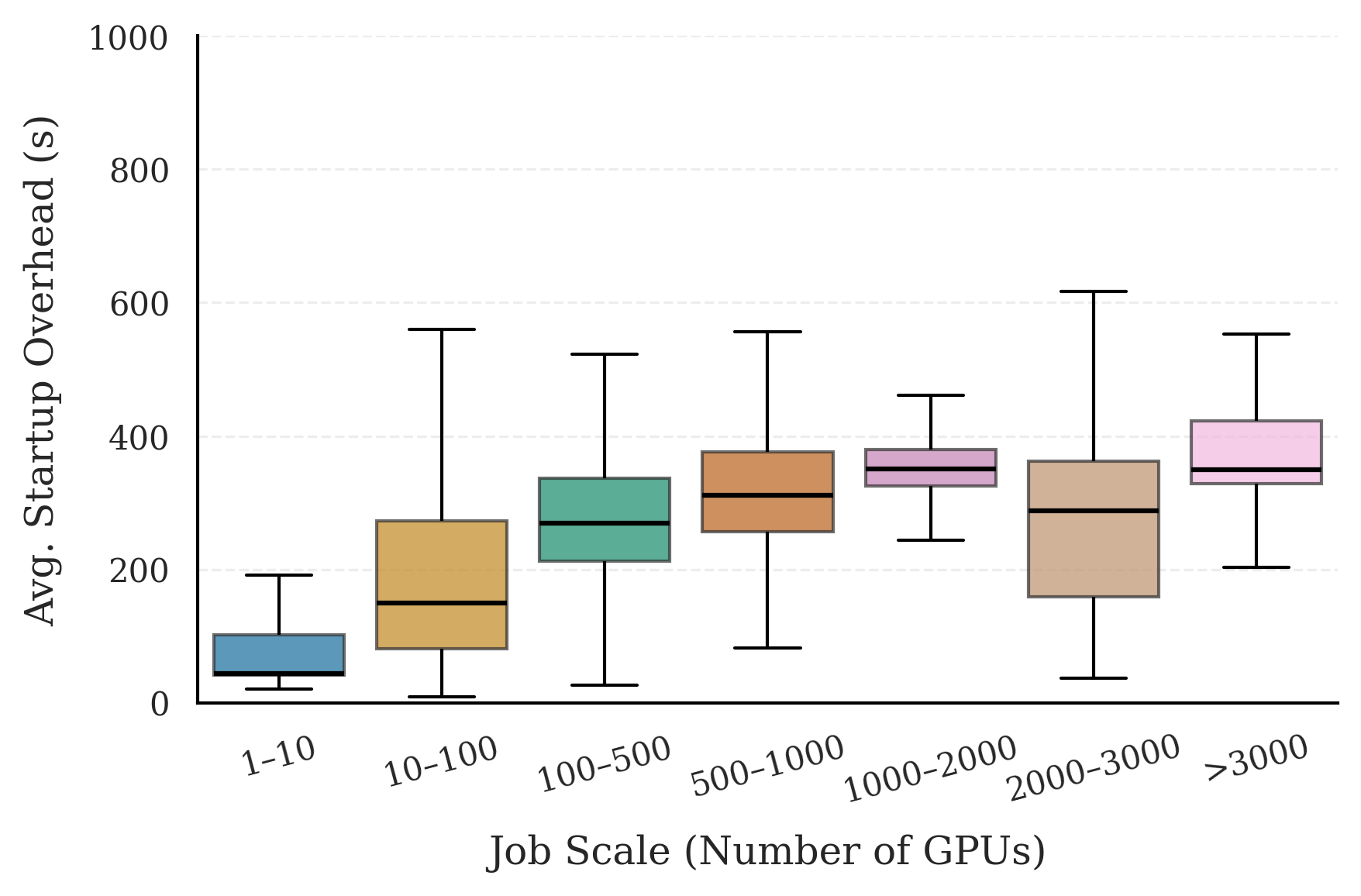}
    \caption{Node-level startup overhead.}
    \label{fig:pod_avg_startup_overhead}
  \end{subfigure}
  \caption{Startup impact of LLM training on job-level (left) and node-level (right) startup overheads. Box plot whiskers extend to two standard deviations, in order to exclude outliers.}
  \label{fig:Impact_of_Startup}
\end{figure*}

\subsection{Opportunities for Optimizing Training Startup}
\label{sec:back_system}
Training large language models (LLMs) is computationally intensive due to the massive scale of both the models and their training datasets~\cite{jiang2024megascale}. 
To handle this complexity, parallelism strategies --- such as data parallelism, pipeline parallelism, and tensor parallelism --- are commonly employed to distribute the workload across multiple GPU devices.
As a result, the training process also becomes communication-intensive.
This computational and communication intensity introduces new challenges and opportunities for system design.

On one hand, LLM training jobs are provisioned with abundant resources. They typically run on dedicated machines, fully utilizing CPU, memory, and GPUs. To support intensive communication, nodes use high-speed interconnects like RDMA and NVLink. Large datasets also demand fast storage systems such as HDFS.

On the other hand, these jobs are complex and require careful coordination for efficiency and stability. Pre-run tests and health checks help prevent failures. During training, monitoring and profiling tools track performance. Many stages require synchronized GPU nodes, where a single straggler can stall the entire job.

These characteristics of LLM training systems present both challenges and opportunities for optimizing LLM startup processes.
First, the abundant computing and communication resources are often underutilized during the startup phase, which creates an opportunity to leverage them more effectively at this stage.
However, the complexity of training jobs makes it difficult to apply some well known techniques, such as those used in serverless startup frameworks.
Furthermore, the straggler problem causes startup overhead to scale with job size, as delays in any single node can impact the entire training job.

In this paper, BootSeer primarily focuses on the three stages in the \textit{Worker Phase}, as these directly waste GPU resources and prone to straggler issues.
Stages in the \textit{Scheduler Phase} do not waste GPU resources because GPU nodes has not been allocated during these stages.
We provide empirical evidence to support this observation in Section~\ref{sec:motivation}.
\section{Characterizing Startup Overhead in LLM Training at Our Large Organization}
\label{sec:motivation}
This section presents a detailed analysis of startup overhead in LLM training, emphasizing its impact on resource efficiency, scalability, and system stability in large-scale production environments.

Note that this section reports \emph{only on data collected using BootSeer/Profiler}, which is described in Section~\ref{sec:method_profilling}.
The capability of BootSeer for optimizing the startup phase was not used here.
Note that this datacenter of the organization is divided into many clusters, each with its own cluster identifier.

The data reported here was taken on a \emph{single cluster} running BootSeer/Profiler over a one-week period.
The cluster comprises a heterogeneous mix of mainstream GPU cards, including NVIDIA H800, A100-80GB, Tesla V100, etc.
Each individual cluster has, in turn, many nodes with 8 or 16 GPUs.
Statistics on \emph{all} jobs submitted to that cluster are reported in this section.
Data collection encompassed over 28,000 submitted jobs, which collectively requested more than 700,000 GPU cards.

\subsection{Impact of Startup Overhead on GPU Utilization and Training Efficiency}
\label{sec:mot_impact}
The first question we address in this section is: \textbf{Why is startup overhead in LLM training important?}
We begin by quantifying how startup overhead contributes to the underutilization of GPU resources and how this inefficiency scales with increasing job size.

To better understand training \textbf{startup overhead} behavior, we analyze it from both the job-level and node-level perspectives.
\begin{itemize}
\item \textbf{Job-level:} Measured from the time a job is submitted to the time training begins.
\item \textbf{Node-level:} Includes the durations of all startup stages, excluding the time spent waiting for other nodes to become ready.
\end{itemize}
Note that node names are assigned at job submission time, but the actual resources are not allocated immediately, which is why the node-level overhead also includes the Resource Queuing and Resource Allocation stages.

Job-level startup overhead increases with job size, leading to significant delays in large-scale training jobs.
Figure~\ref{fig:trial_avg_startup_overhead} shows the job-level average startup overhead for different ranges of job scale.
In our cluster, large training jobs requiring more than 100 GPU cards typically take around 6–-7 minutes to start.
In the worst case, some jobs require 15~minutes or more before training begins.
Smaller jobs tend to start more quickly, as they typically involve smaller container images and smaller model checkpoints.

Larger jobs are more prone to repeated startups, which increases GPU waste due to debugging and reconfiguration efforts.
Figure~\ref{fig:pod_avg_startup_overhead} shows the node-level startup overhead.
Compared to Figure~\ref{fig:trial_avg_startup_overhead}, the node-level overhead exhibits a similar trend: smaller jobs start more quickly.
Additionally, node-level startup overhead is consistently lower than job-level overhead:  by approximately one minute for the same job scale.
This difference arises because of the straggler effect, which delays overall job readiness, although it does not affect the accuracy of node-level measurements.

\begin{figure}[t]
  \centering
  \includegraphics[width=0.75\linewidth]{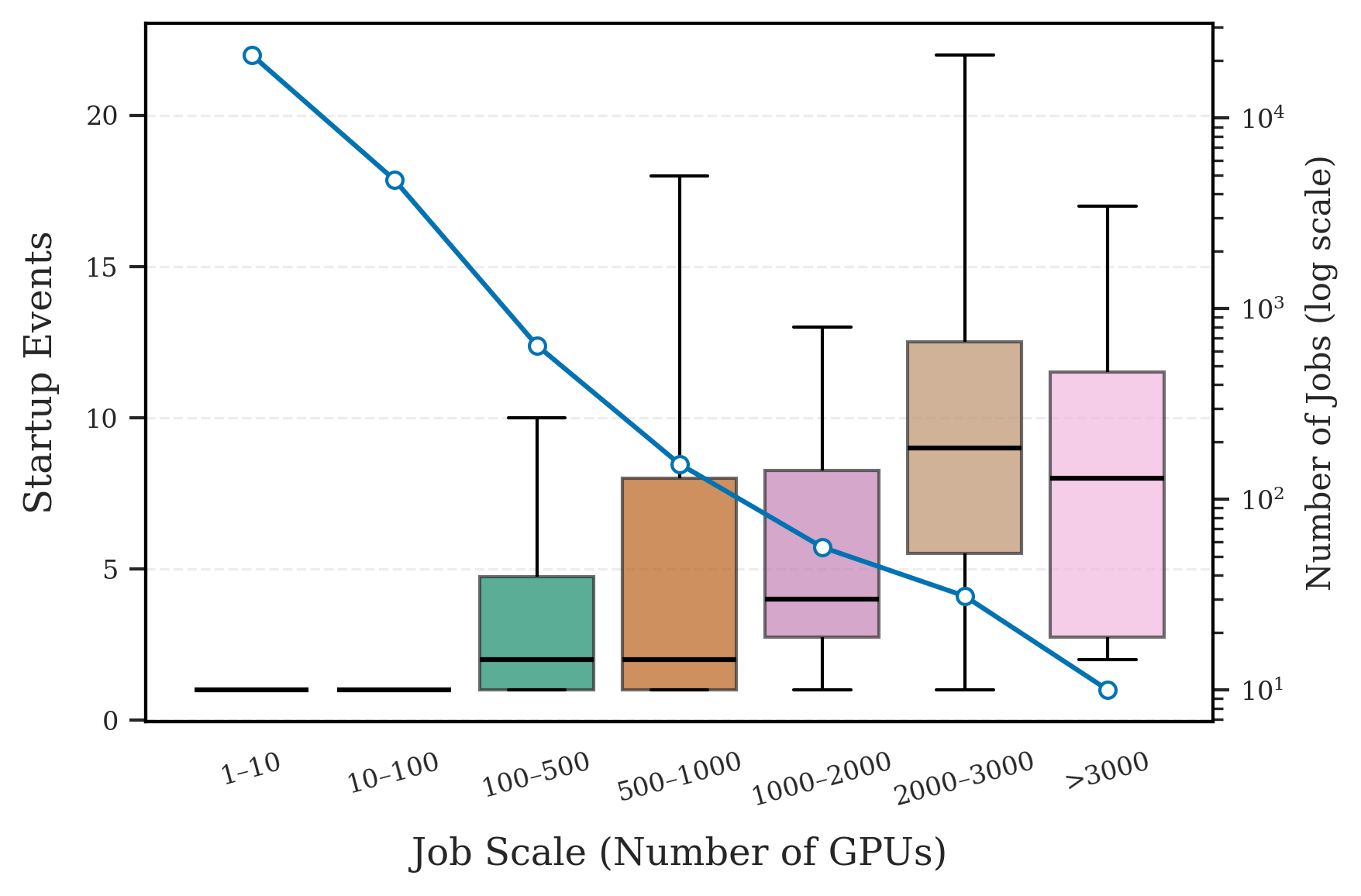}
  \caption{Number of startup events per job (left y-axis) and number of jobs (right y-axis). Box plot whiskers extend to two standard deviations, in order to exclude outliers.}
  \label{fig:trial_startup_count}
\end{figure}

Larger jobs tend to experience multiple startups, amplifying GPU waste and highlighting inefficiencies in debugging and configuration workflows.
Figure~\ref{fig:trial_startup_count} presents the number of startup events per job and the number of jobs \hbox{vs.} a job scale range.
Smaller jobs, those using fewer than 100 GPUs, typically experience only a single startup during their life cycle.
In contrast, larger jobs undergo multiple startups, ranging from 2 to 8 startups in most cases. 
In the worst case, some jobs may trigger 20 or more startups, often due to debugging or repeated configuration changes.
Unfortunately, because larger jobs request more GPUs, each startup results in a greater amount of wasted GPU time.

\textbf{Key Takeaway: Startup overhead is non-negligible, and its impact increases with job scale.}

\subsection{Component-level Breakdown of Startup Overhead}
\label{sec:mot_breakdown}
To understand where time is spent during startup, we break down the node-level startup overhead into distinct stages, ranging from Resource Queuing to Model Initialization.
In our system, physical resources are not actually allocated during the Resource Queuing and Resource Allocation stages; time spent in these stages only affects the user experience and the efficiency of the debug-resubmit cycle.
In contrast, the Image Loading, Environment Setup, and Model Initialization stages actively consume GPU resources and thus represent true overhead.

\begin{figure}[t]
  \centering
  \includegraphics[width=0.65\columnwidth]{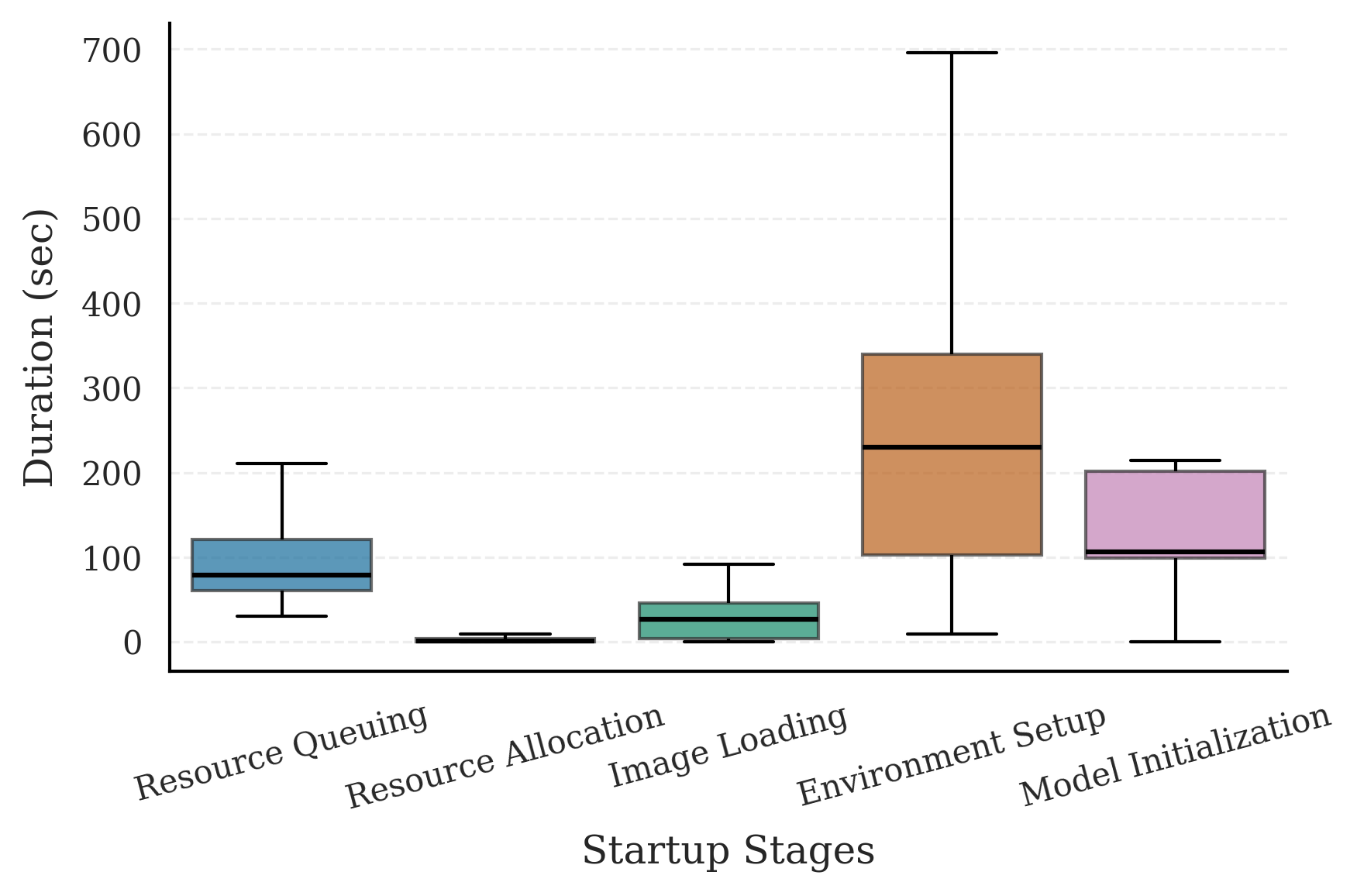}
  \caption{Breakdown of node-level startup overhead across different initialization stages. Box plot whiskers extend to two standard deviations, in order to exclude outliers.}
  \label{fig:break_down_startup}
\end{figure}

Non-GPU-consuming stages, especially Resource Queuing, can introduce significant delays even before actual resource usage begins.
Figure~\ref{fig:break_down_startup} illustrates the components of node-level startup overhead across various initialization stages.
For the stages that do not consume GPU resources, a job typically waits around 100 seconds for sufficient resources to become available, although in extreme cases, the wait time can extend to several hours.
The Resource Allocation stage, by contrast, is trivial and usually completes within a few seconds.

Among the GPU-consuming stages, startup overhead is mainly driven by Image Loading, Environment Setup, and Model Initialization.

The \textbf{Image Loading stage} performs as expected, as our system already employs lazy loading of data blocks.
This stage typically takes 20 to 40 seconds to download the image and start the container.

The \textbf{Environment Setup stage} is the most significant bottleneck in our system.
It takes 100 to 300 seconds to complete, and its variability can also lead to potential stragglers.
As discussed in Section~\ref{sec:back_startup}, we install certain dependencies during this stage primarily for two reasons:
\begin{enumerate}
    \item some packages have multiple versions, and the correct version is determined only at runtime; and
    \item some packages are frequently updated, and so the package version frequently changes.
\end{enumerate}
However, this installation process is time-consuming and increases variability in node-level duration, making it the primary contributor to a longer job-level duration.

The \textbf{Model Initialization stage} is the second major bottleneck, taking 100 to 200 seconds to initialize the training program and resume from checkpoints.
Checkpoint files are usually stored in an HDFS cluster, and the training nodes must access them over the network—introducing another potential straggler reason.

\textbf{Key Takeaway: GPU resource waste (idle time and initialization) is dominated by startup, which in turn is dominated by Environment Setup, Model Initialization, and image loading.  Environment Setup is the largest of the  bottlenecks, due to the high variability of the time for installing runtime dependencies.}

\subsection{Startup Stragglers}
\label{sec:mot_straggler}
To address how startup stragglers emerge and become more of a bottleneck as job scale increases, we examine the relationship between startup delays and the number of machines involved. 
Stragglers typically occur during stages where worker nodes must access remote resources over the network. 
Common examples include downloading images, installing dependencies during the Environment Setup stage, and retrieving checkpoint files from the HDFS filesystem.

In this section, we examine the execution time for the dependency installation script that is run during the Environment Setup stage.  This execution time serves as a proxy to measure the straggler effect during startup.

We introduce a metric, \emph{the Max/Median Ratio}, to quantify the severity of node-level startup stragglers within each job.
Specifically, the Max/Median Ratio is the startup time of the slowest node divided by the startup time of the median node within the same job, using the execution time of the dependency installation script as a proxy for startup time.
This ratio reflects how long the rest of the nodes must wait for the worst straggler to complete.

\begin{figure}[htbp]
  \centering
    \includegraphics[width=0.65\linewidth]{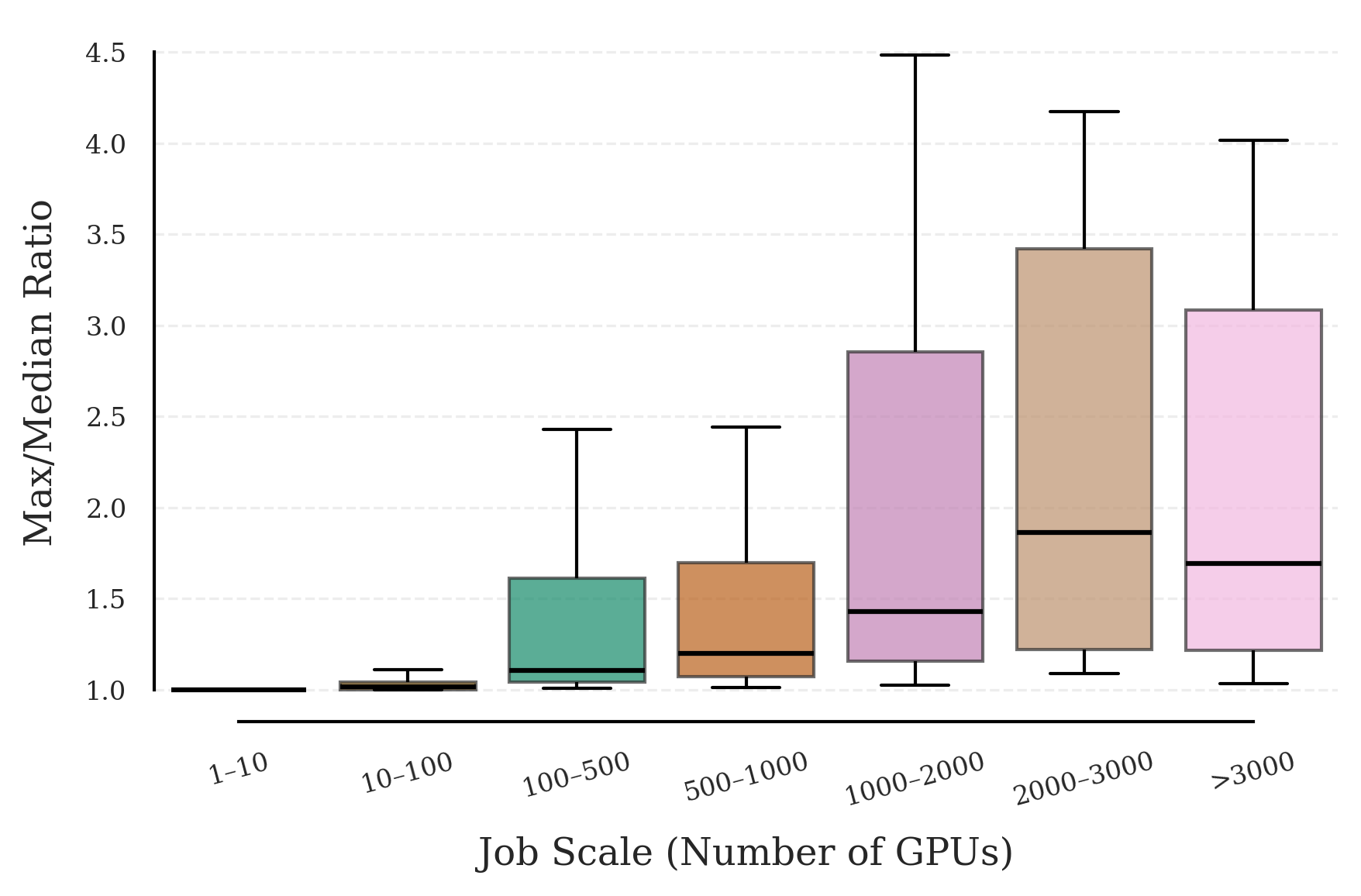}
    \caption{The straggler effect increases with job scale. The y-axis shows the Max/Median ratio for each of a range of job scales. Box plot whiskers extend to two standard deviations in order to exclude outliers.}
    \label{fig:straggler_effect_vs_gpu_scale}
\end{figure}

The straggler effect worsens significantly with job scale, leading to longer delays in large GPU jobs.
Figure~\ref{fig:straggler_effect_vs_gpu_scale} illustrates the relationship between the straggler effect and job scale. When the job scale is small, the straggler effect is minimal. 
However, the Max/Median ratio increases noticeably as the number of GPUs grows. 
For jobs running on more than 1,000 GPUs, stragglers can slow down the dependency installation duration by approximately 1.5~times. 
In extreme cases, the slowdown can reach 4~times or more.

\begin{figure}[htbp]
  \centering
    \includegraphics[width=0.65\linewidth]{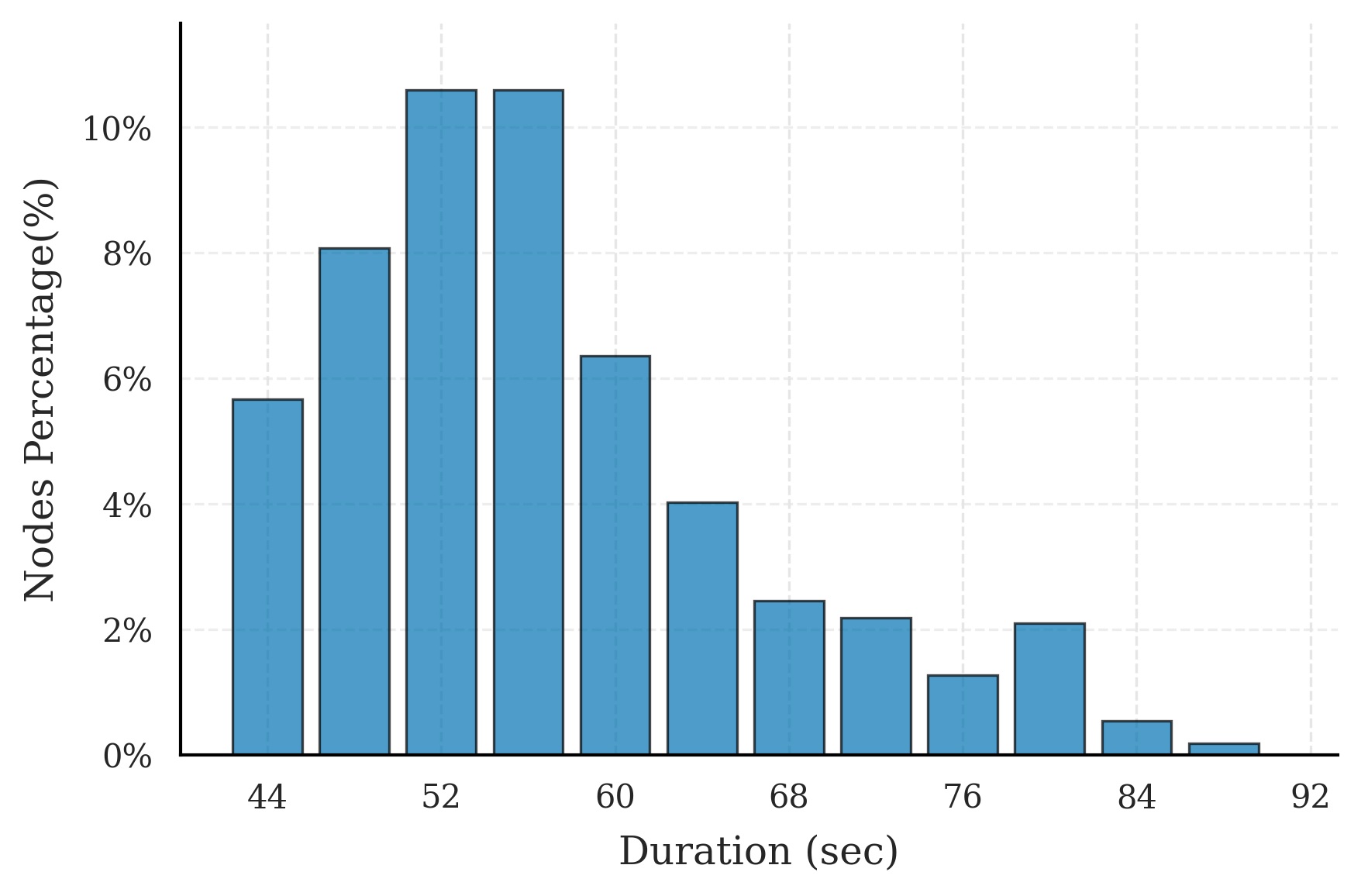}
    \caption{Distribution of dependency installation script durations for a job running on 1,440 8-GPU servers (one of the largest jobs in the trace).}
    \label{fig:straggler_install_dependency}
\end{figure}

In large-scale jobs, even a rare slow node can end up stalling thousands of GPUs.
Figure~\ref{fig:straggler_install_dependency} presents the distribution of worker nodes versus the duration of dependency installation for one of the largest job in our trace, which involves 11,520 GPUs. 
The distribution exhibits a clear long-tail pattern: most machines complete the installation within 60 seconds, while fewer than 1\% of nodes take as long as 92 seconds. Unfortunately, all 1440 servers must wait for the slowest node to finish before proceeding.

\textbf{Takeaway: The straggler effect during startup, typically as worker nodes access remote resources, is especially detrimental to large-scale training jobs.}

\subsection{Case Study: Startup Stragglers and Stability}
\label{sec:mot_stability}
Through real-world examples, we demonstrate how instability during the startup phase can lead to job failures, slowdowns, and jobs freezing within production environments.
The data shown here is manually selected from system logs (not from BootSeer/Profiler) to illustrate potential system instability in jobs involving very many nodes.

\textbf{Startup Slowdown in a Multimodal Model Training with 11,520 Nvidia H800 GPUs}
We analyzed a large-scale multimodal model training job deployed on 1,440 nodes, each equipped with 8 GPUs. During startup, a significant performance anomaly was observed in the data preparation phase.
Specifically, during the Environment Setup stage, worker nodes need to install an NCCL related package. While most nodes completed the download and extraction in approximately 6 seconds, a subset experienced severe delays of up to 90 seconds.
Further investigation identified the root cause as throttling due to high-concurrency access: over 1,000 nodes performed simultaneous pull operations, overwhelming the source control management (SCM) backend and triggering rate limiting. This case illustrates how large-scale parallelism can induce unexpected bottlenecks in external package distribution systems, adversely affecting startup efficiency.

\textbf{Startup Failure During a Model Training on 2,016 Nvidia H800 GPUs}
In another instance, we launched a training job using 252 nodes, each with 8 GPUs. During initialization, the system attempted to install an internal Python package.
However, due to high-concurrency access across all nodes, the installation was throttled, leading to download failures. These failures triggered errors that caused the entire job to terminate during startup.
This case highlights the fragility of large-scale startup procedures, where centralized package distribution systems can become single points of failure. It underscores the importance of robust dependency pre-distribution or caching strategies to maintain stability under high-parallelism conditions.
\section{BootSeer Design}
\label{sec:design}
This section introduces the full \textbf{BootSeer} (both profiler and startup optimizer), a framework designed to systematically analyze and optimize startup overhead in large language model (LLM) training. 

Using the profiling system described in Section~\ref{sec:method_profilling}, we identify three major bottlenecks in the startup process: container image loading, runtime dependency installation, and checkpoint resumption. These stages are not only significant sources of latency, but also common causes of straggler effects that delay overall job startup.

To address these challenges, BootSeer applies targeted optimizations: Section~\ref{sec:method_image-loading} presents a lazy and peer-assisted image loading strategy; Section~\ref{sec:method_env_cache} introduces a job-level environment cache to reduce redundant package installation; and Section~\ref{sec:method_resume_ckpt} describes a striped I/O mechanism for efficient checkpoint resumption.

The description of how these optimizations are integrated into an LLM is novel.
While similar optimizations have been frequently described in other domains,
in the context of large language models (LLMs), these optimizations serve a more specialized purpose: reducing significant bottlenecks during the startup phase.
This is important in two common types of training workloads. 
The first involves long-running training jobs that frequently stop and restart due to failures, upgrades, or debugging. 
The second consists of a large number of shorter, similar jobs used to test new features—typically lasting only a few hours or less.

\subsection{Profiling System for LLM Startup Stages}
\label{sec:method_profilling}
\begin{figure}[htbp]
  \centering
  \includegraphics[width=0.65\columnwidth]{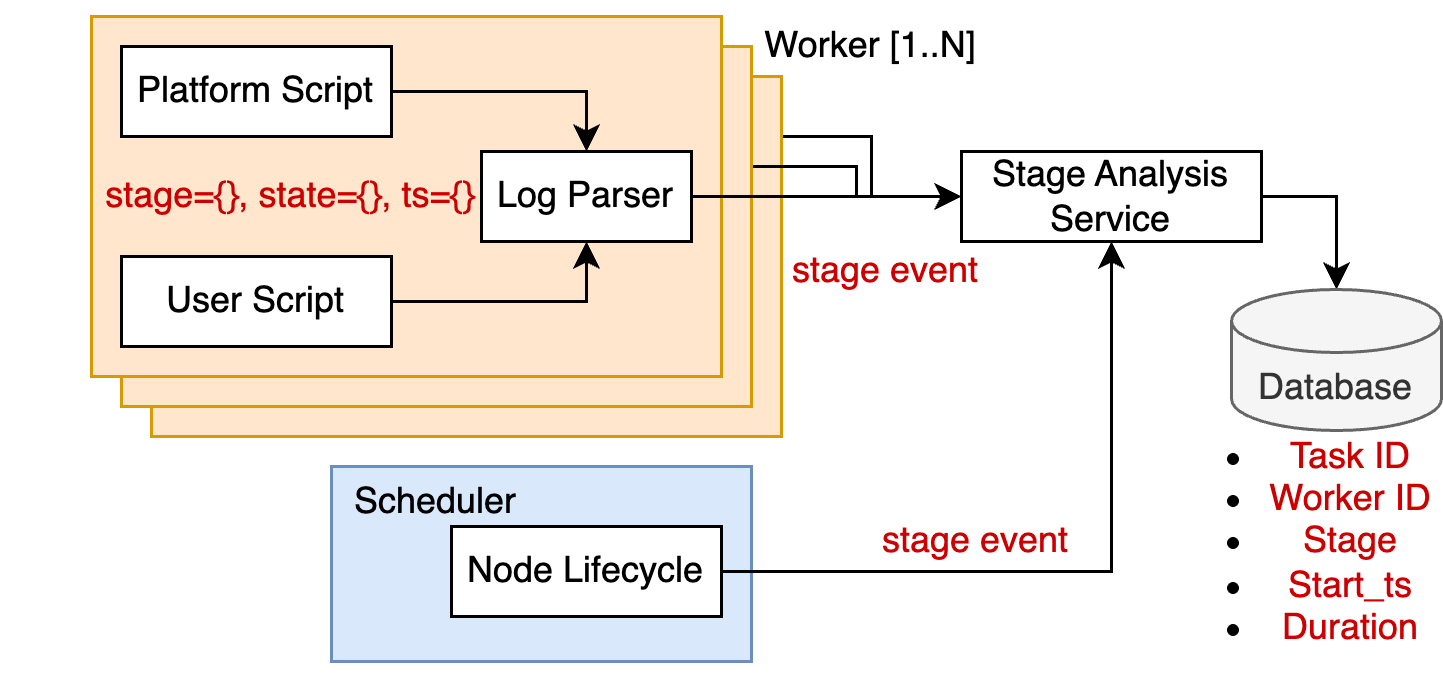}
  \caption{Overview of BootSeer's profiling system for LLM startup stages. Stage transitions are logged on each worker node, parsed locally, and sent to a central Stage Analysis Service for duration computation and visualization. ‘ts’ denotes timestamp.}
  \label{fig:startup_profilling}
\end{figure}

To effectively optimize startup overhead in LLM training, BootSeer first provides detailed measurement of each startup stage through a lightweight profiling system. 
BootSeer implements a lightweight event collection and analysis system for profiling these stages, as illustrated in Figure~\ref{fig:startup_profilling}.

BootSeer implements a lightweight logging system by inserting `print` or `echo` commands to record specific stage transitions during startup.  
Each worker node runs a local \textit{Log Parser} that extracts stage events from these log files and sends them to a centralized \textit{Stage Analysis Service}.  
In addition, \textit{Node Lifecycle} events, such as queuing for resources or pulling container images, are also recorded and forwarded to the Stage Analysis Service.  
After grouping and analyzing these stage events, the Stage Analysis Service stores the computed stage durations in a database for further analysis and visualization.

\subsection{Prefetching-Based Image Loading Optimization}
\label{sec:method_image-loading}

The key challenge in this stage is the I/O and network bandwidth bottleneck, as the container images used in LLM training are relatively large—typically exceeding 25 GB.  
If all worker nodes attempt to download these images concurrently from a remote source, it can create severe pressure on the network and I/O subsystems, leading to unacceptable startup delays and frequent straggler effects.

\textbf{Baseline:} Our baseline system builds on recent advances in lazy loading for large container images to address the image loading bottleneck.
As shown in prior work on optimizing image loading~\cite{brooker2023demand, liu2025flacio}, lazy loading is particularly effective for large container images. Our platform adopts block-level on-demand loading, lazy loading, and block deduplication. Instead of using traditional Open Container Initiative (OCI) images with a layered filesystem, we flatten all layers into a single unified layer. The image contents are managed at the block level, enabling both efficient deduplication and runtime lazy loading. This approach achieves up to a 10× improvement over traditional OCI.

\textbf{BootSeer Optimizations:} To further accelerate image loading beyond baseline lazy loading, BootSeer introduces two new optimizations:
(1)~record-and-prefetch of hot blocks; and (2)~peer-to-peer sharing of data blocks.

\begin{figure}[htbp]
  \centering
  \includegraphics[width=0.55\linewidth]{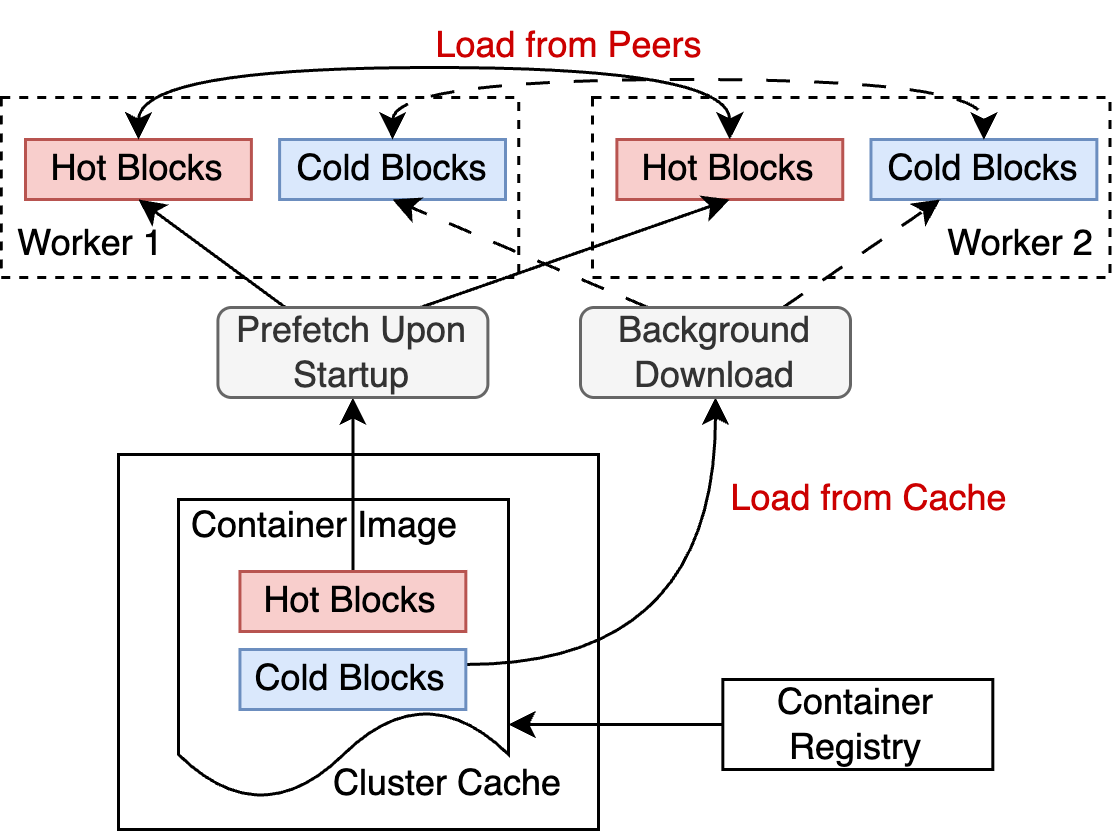}
  \caption{Image loading optimization in BootSeer: hot blocks are recorded and prefetched, then transferred peer-to-peer.}
  \label{fig:prefetch_container}
\end{figure}

The key insight behind BootSeer's optimization is that \textit{training images are sparsely accessed during startup}.
Consistent with observations from prior work~\cite{harter2016slacker}, only a small subset of image data is accessed in the early stages of training.
This compact and stable set of blocks presents an ideal opportunity for prefetching.
Figure~\ref{fig:prefetch_container} illustrates the record-and-prefetch mechanism employed by BootSeer.
During the first access to a container image, BootSeer identifies and records the “hot” blocks accessed during startup.
In subsequent runs using the same image, these hot blocks are proactively prefetched to accelerate container startup.

\textbf{The record phase} captures access patterns during container startup to identify hot data blocks.
The container runtime on the worker node captures the absolute file paths and block offsets of the data blocks accessed during startup.  
This access trace is then uploaded to a remote service. 
When the same image is used by another worker node, its container runtime can retrieve this record from the controller to identify and prefetch the hot blocks.

\textbf{The prefetch phase} uses prior access records to proactively load hot blocks, reducing startup delays.
During image loading, the container runtime prefetches all identified hot blocks to ensure fast startup.  
After the container begins execution, the runtime continues downloading the remaining cold blocks in the background to the worker node, thereby avoiding remote fetches during training.  
Notably, all data blocks can be retrieved either from a cluster-level cache or from peer nodes.  
Since LLM training jobs typically involve multiple machines pulling the same image concurrently, fetching from a single source can impose significant pressure on network bandwidth.  
Leveraging peer-to-peer sharing helps distribute this bandwidth load across multiple links and reduces the risk of straggler nodes.

\subsection{Environment Caching for Job-Level Dependency Installation}
\label{sec:method_env_cache}
The Environment Setup phase, as introduced in Section~\ref{sec:back_startup}, includes dependency installation and daemon initialization, which are critical for preparing the runtime environment.
\textbf{BootSeer installs package dependencies at this phase instead of placing all dependency packages in the Dockerfile because:}
(1)~some dependency installations are decided based on runtime environment (e.g. GPU type, OS version); and
(2)~there are trade-off between flexibility and efficiency:  some dependencies are upgraded very frequently and frequently building new container for new dependency versions would usually be time-consuming.

\textbf{Baseline:} In the baseline system, dependencies are installed on-the-fly using commands such as pip install, but this approach introduces several scalability bottlenecks.
And all training machines install a specific dependency at the same time, producing a "bit storm", i.e., bringing pressure on the network to deliver the new package.
As the pressure on the network builds up, the network may limit traffic from the package source.  The result is one or more straggler nodes.

\textbf{BootSeer Optimizations:} To improve the efficiency of package installation while preserving flexibility, BootSeer uses a job-level environment cache to skip redundant dependency installations. 
This cache is generated during the first run of a job and is reused in subsequent runs, restarts, or node replacements for the same job. 
If the job parameters change, the cache is marked as expired.
\begin{figure}[t]
  \centering
  \includegraphics[width=0.55\columnwidth]{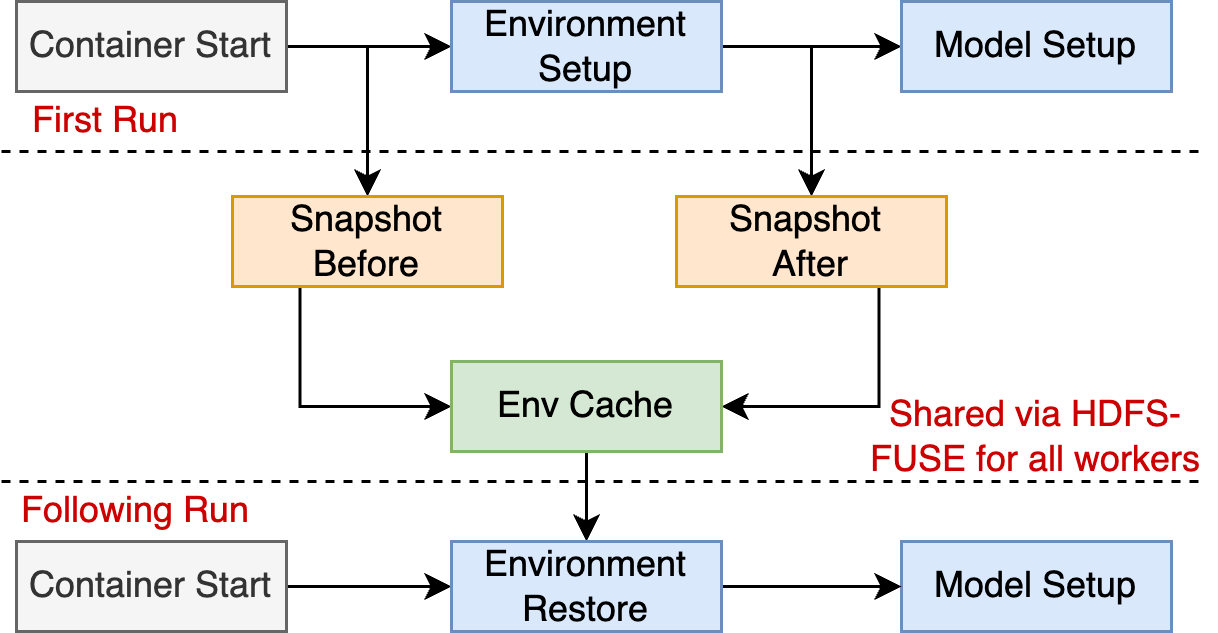}
  \caption{Job-Level Environment Cache}
  \label{fig:depedency_snapshot}
  \vspace{-3mm}
\end{figure}

The environment cache is generated by capturing file system changes during the initial environment setup.
Figure~\ref{fig:depedency_snapshot} shows the process of creating and using the environment cache.
In the first run of a job, BootSeer create the environment cache based on the difference of \textit{Target Directory} before and after the \textbf{Environment Setup} phase on worker node 0,
where the Target Directory is the dependency install path, e.g., the site-package directory.
All files added or modified during the first environment setup are compressed and uploaded to HDFS through HDFS-FUSE.

Subsequent runs of the same job leverage the saved cache to skip repeated installations, significantly reducing startup time. 
If the cache exists, BootSeer restores added and modified files through HDFS-FUSE and skips all install commands. 
If runtime parameters change (e.g., dependency versions or GPU type), BootSeer marks the cache as expired and creates a new one.

By using Environment Cache, BootSeer reduces the startup overhead but still supports the necessary flexibility.
BootSeer shares the environment among all training machines with HDFS-FUSE, thus reducing bandwidth overhead during startup.
The HDFS-FUSE filesystem is introduced in Section~\ref{sec:method_resume_ckpt}
Eliminating duplicate install commands and unnecessary network traffic also helps to avoid straggler effects during startup.

\subsection{Striped HDFS-FUSE for Efficient Checkpoint Resumption}
\label{sec:method_resume_ckpt}
During the \textbf{Model Initialization} stage, checkpoint resumption is the only phase that requires access to remote resources.
LLM checkpoints are relatively large files.
For example, checkpointing of an MOE model with 25~billion parameters can result in a 400~GB checkpoint file.
On our platform, checkpoint images are stored in an HDFS cluster to ensure high I/O performance and necessary data redundancy.
However, concurrently downloading these large checkpoint files to worker nodes places significant pressure on I/O resources and increases the risk of more straggler nodes.
To address this, BootSeer implements a striped parallel read mechanism to accelerate the checkpoint resumption process.

\textbf{Baseline:} Our baseline system stores checkpoints in HDFS and mounts them via HDFS, but this method lacks the I/O parallelism needed for fast checkpoint resumption.
Training program need to download checkpoint files from HDFS, and resume the parameters. 
We also implement HDFS-FUSE~\cite{vangoor2017fuse}, so that the remote HDFS directory can be mounted to a local directory.

\begin{figure}[h]
  \centering
  \includegraphics[width=0.65\columnwidth]{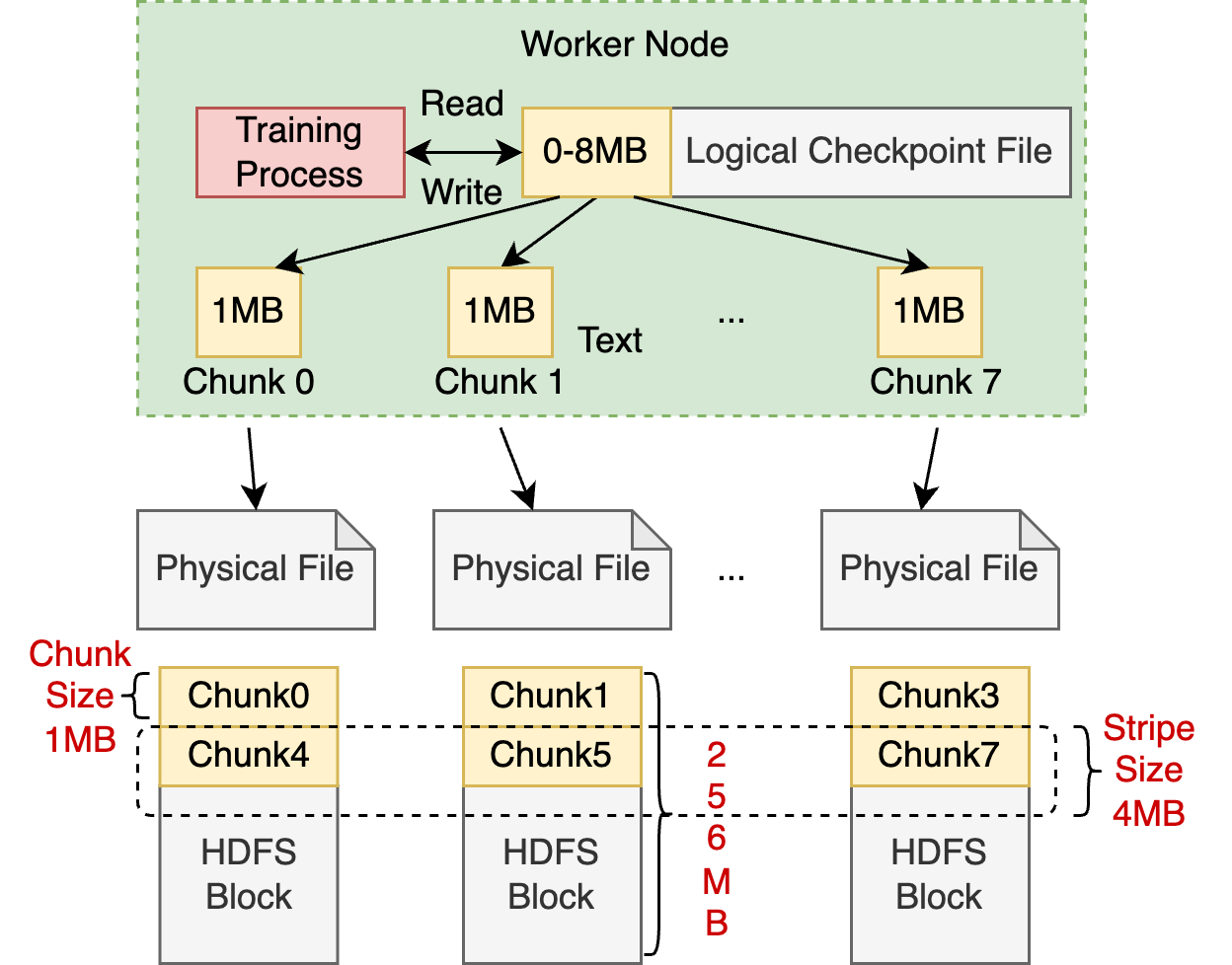}
  \caption{Striped HDFS-FUSE layout. The logical checkpoint file is split into 1MB chunks, written to physical files, and stored in HDFS using 4MB stripes. Chunks from multiple files are interleaved within each HDFS block.}
  \label{fig:striped_hdfs_fuse}
\end{figure}

\textbf{BootSeer Optimizations:} To overcome the I/O bottlenecks of baseline HDFS usage, BootSeer introduces a striped I/O mechanism for checkpoint access.
Given the ample compute resources available on training nodes during the startup stage, the worker node have redundant resources to download data chunks in parallel.
Figure~\ref{fig:striped_hdfs_fuse} illustrates the layout of the striped HDFS-FUSE system.

In the original HDFS design, data is written sequentially in large blocks (typically 512 MB), with each block stored within a single replication group of DataNodes.
This architecture limits I/O parallelism during both read and write operations, which can become a significant bottleneck when accessing large model checkpoints.

Striping addresses this limitation by dividing data into smaller chunks and distributing them across multiple DataNode groups.
This enables high parallelism during read operations, allowing different parts of a large file to be accessed simultaneously.
As a result, checkpoint resumption is faster compared to the original download-and-resume approach.

\section{Evaluation}
\label{sec:eval}
\subsection{Experiment Setup}
\label{sec:eval_setup}
In this section, we introduce the workload, platform, evaluation metrics, and research questions as follows:

\textbf{Training Workload.} We use large-scale MoE training jobs with significant checkpoint and container sizes to evaluate startup overhead under realistic, heavy-load conditions
Each job trains an 8-layer Mixture-of-Experts (MoE) model with 128 experts per layer. 
The model is configured with 2-way pipeline parallelism, and the checkpoint size is 413~GB. 
We scale the training across 16, 32, 48, 64, 128, 512 and 1024 GPU cards, corresponding to data parallelism degrees of 1, 2, 3, 4, 8, 16 and 32, respectively. 
The container image size for this job is 28.62~GB. 
Note that the evaluation task we selected is a training job typically used at large scales (1000+ GPUs), and therefore does not align with the statistical characteristics described in Section~\ref{sec:motivation}.

\textbf{Test Platform.} 
Experiments are run on a high-end GPU cluster representative of modern production environments, with nodes equipped with H800 GPUs and InfiniBand interconnects.
Each server is equipped with two Intel$^\textrm{\textregistered}$ Xeon$^\textrm{\textregistered}$ Platinum 8457C CPUs (96 cores per CPU), 2.9~TB of memory, and 8 NVIDIA H800 GPUs. 
We use a total of 128 nodes. 

\textbf{Metric.} Since our focus is on the performance of startup overheads, we exclude the Resource Queuing and Resource Allocation stages, as they introduce external variability. Therefore, in this experiment, we only include the following phases: \textbf{Image Loading}, \textbf{Environment Setup}, and \textbf{Model Initialization}. 
Startup overheads are aggregated at the job level to reflect the end-to-end startup time of each training job.
We refer to the ``Baseline'' as the original platform without the benefit of BootSeer.

\textbf{Our study focuses on three key research questions to evaluate the effectiveness and granularity of BootSeer’s optimizations.}
\begin{itemize}
    \item \textbf{RQ1:} How does BootSeer perform compared to the baseline in terms of end-to-end startup overhead?
    \item \textbf{RQ2:} What is the optimization effect of BootSeer on each startup stage?
    \item \textbf{RQ3:} How does BootSeer impact the straggler effect compared to the baseline?
\end{itemize}

\subsection{End-to-End Startup Optimization}
\label{eval:end_to_end}
To answer \textbf{RQ1}, we conduct an end-to-end experiment to evaluate the optimization performance of BootSeer compared to the baseline.
Both frameworks are used to start the MoE training job described in Section~\ref{sec:eval_setup}. All results are averaged over three independent experiments.
Before each run, we clear the local image cache to eliminate potential impacts.

To explain the performance gap, we detail the startup mechanisms used in both the baseline and BootSeer.

\textbf{Baseline.} The baseline follows our original training startup process: (1) the container image is downloaded using a lazy-loading mechanism with peer-to-peer sharing across all worker nodes, (2) dependencies are installed during startup, and (3) checkpoint files are retrieved from HDFS and loaded into memory to resume training.

\textbf{BootSeer.} The BootSeer setup includes the following components: 
\textbf{(1) Record-and-prefetch for container images} — Hot data blocks are identified based on access patterns from previous runs using the same container image. A two-minute window is used to capture hot blocks, while cold blocks are prefetched in the background using eight threads. 
\textbf{(2) Environment cache} — Environment caching is enabled, with cache files generated from previous executions of the same task. The compressed cache (270~MB) is stored in HDFS and restored during environment setup. 
\textbf{(3) Checkpoint resumption} — Checkpoint files are mounted to a local directory using striped HDFS-FUSE during the setup phase. HDFS-FUSE is supported by an auxiliary container that must also be pulled during image loading.

\begin{figure}[h]
  \centering
  \includegraphics[width=0.85\columnwidth]{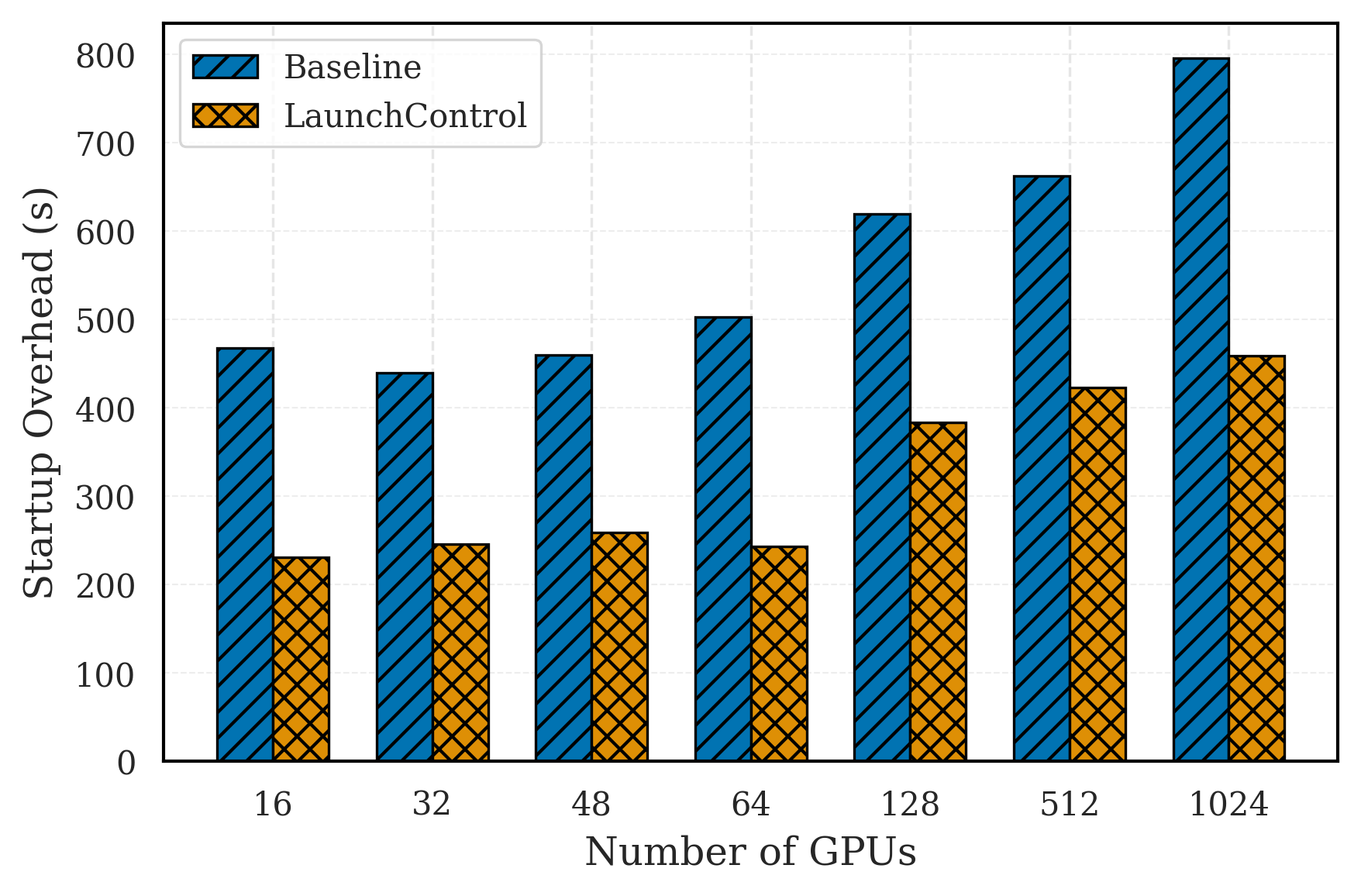}
  \caption{End-to-end startup overhead comparison}
  \label{fig:end-to-end}
\end{figure}

\textbf{BootSeer consistently reduces end-to-end startup time by 2 times compared to the baseline, and this benefit holds across different job scales.}
Figure~\ref{fig:end-to-end} presents the startup overhead of both frameworks with respect to the scale of the training jobs.
The startup overhead increases with the size of the training cluster, especially when scaling from 64 to 1024 GPUs. This trend aligns with the findings discussed in Section~\ref{sec:motivation}.

\subsection{Improvement Breakdown}
\label{eval:eval_breakdown}
\begin{figure}[ht]
  \centering
  \includegraphics[width=\columnwidth]{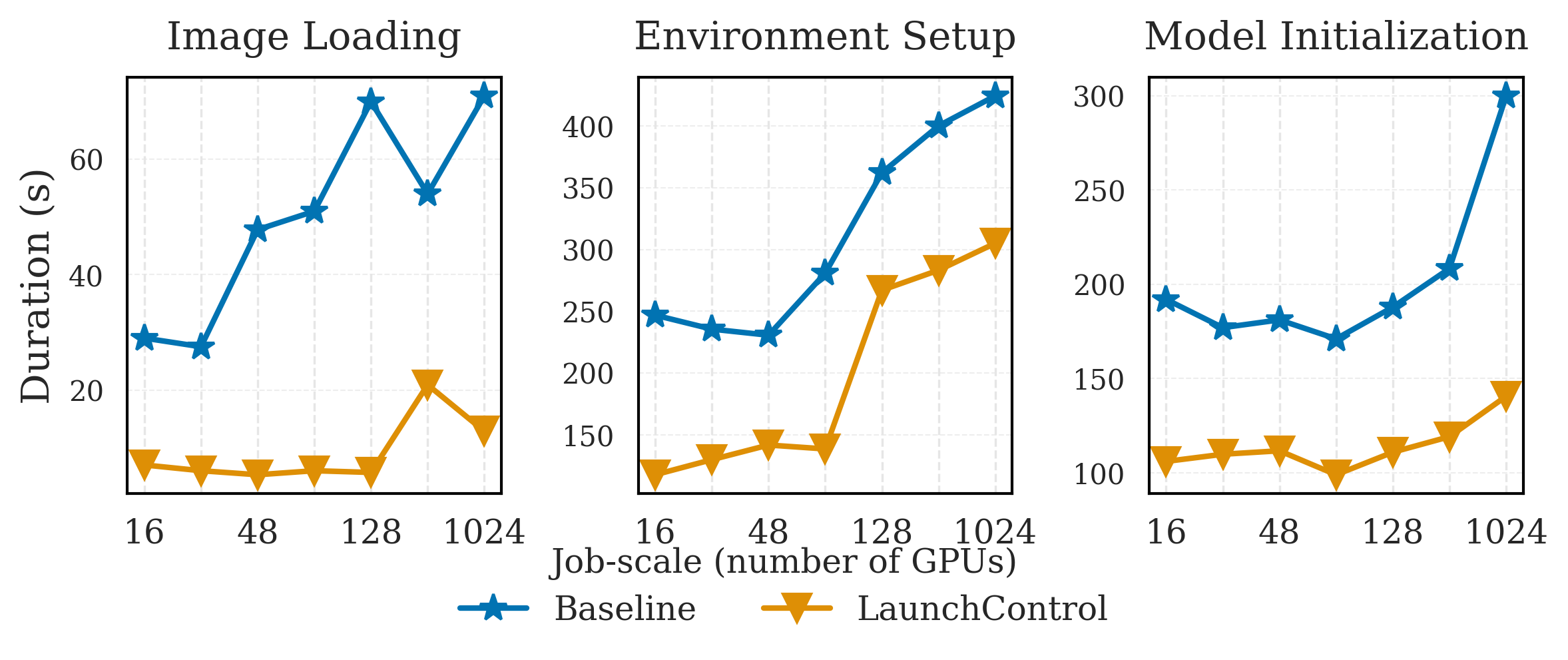}
  \caption{Breakdown of training startup overhead for BootSeer and the baseline. Each plot shows the duration of individual startup stages, with the x-axis representing the number of GPUs.}
  \label{fig:breakdown_improvement}
  \vspace{-3mm}
\end{figure}

To address \textbf{RQ2}, we break down the startup overhead into three stages: Image Loading, Environment Setup, and Model Initialization.
Figure~\ref{fig:breakdown_improvement} shows the duration of each stage across different job scales.

\begin{figure*}[tbp]
  \centering
  \includegraphics[width=\columnwidth]{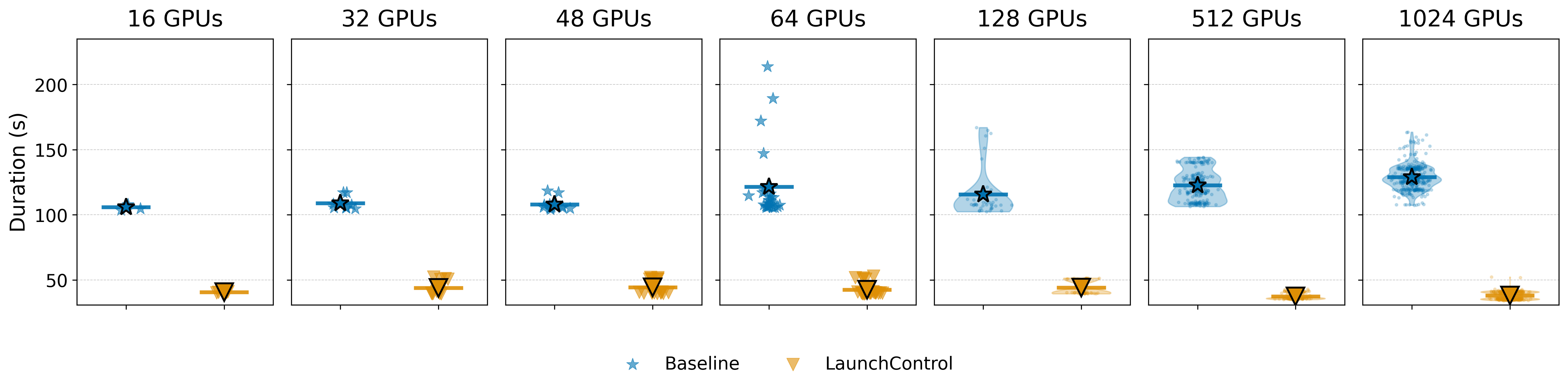}
  \caption{Distribution of execution times for the dependency installation script (a component of the Environment Setup stage) across varying job scales.}
  \label{fig:straggler_exp}
\end{figure*}

\textbf{Image Loading.} BootSeer outperforms the baseline by a factor of 4× to 7× as job scale increases.
This improvement stems from inefficiencies in the baseline's lazy-loading mechanism: when a container accesses a data block that has not yet been loaded onto the local node, it must retrieve the block from a remote cache or another machine. As the number of GPUs increases, such cache misses place additional strain on the cluster's network. In contrast, BootSeer’s record-and-prefetch strategy eliminates remote data access, maintaining stable loading times regardless of scale.

\textbf{Environment Setup.} BootSeer reduces latency by 20\% to 50\% by bypassing most dependency installation commands.
However, as the job scale increases from 128 to 1024 GPUs, we observe a significant rise in duration and diminishing performance gains. This trend is attributed to the growing overhead of establishing and synchronizing connections across all nodes, a process that BootSeer does not explicitly optimize.

\textbf{Model Initialization.} BootSeer reduces overhead by up to 2.6× through more efficient checkpoint resumption.
Among all stages, Model Initialization shows the most stable performance, with no substantial increase in duration as job scale grows. This stability may be due to the current job sizes not being large enough to cause bottlenecks in our HDFS.

\subsection{Troubleshooting Startup Stragglers} 
\label{sec:eval:straggler}
To address \textbf{RQ3}, we analyze how our environment cache helps mitigate stragglers in large-scale training.
As repeatedly emphasized throughout this paper, access to remote resources is a major cause of stragglers during training startup.
This observation motivates the design of a job-level environment cache, aimed at reducing startup overhead while preserving the necessary flexibility.

BootSeer significantly reduces variability in dependency installation times, effectively eliminating startup stragglers.
Figure~\ref{fig:straggler_exp} shows the distribution of execution times for the dependency installation script executed during the Environment Setup stage.
Execution times were collected from each worker node across different job scales.
The results demonstrate that \textbf{BootSeer not only reduces package installation overhead, but also eliminates stragglers}.

The value of environment caching becomes more significant as job scale increases, where straggler effects grow more severe.
Although the straggler effect in this experiment is based on just a few hundred GPUs, the results in Section~\ref{sec:mot_straggler} indicate that the effect becomes more pronounced as the job scales to hundreds or thousands of GPU servers.
\section{Related Work}
\label{sec:related_work}
\textbf{Optimization of LLM Training Systems.} 
A growing body of systems research has focused on improving the efficiency and stability~\cite{narayanan2021efficient, jiang2024megascale, wan2025bytecheckpoint,wu2024falcon, dong2024boosting, cui2025xputimer} of LLM training.
Many recent studies have focused on communication-efficient techniques to accelerate LLM training. 
Examples include the use of gradient compression~\cite{alistarh2017qsgd} and mixed-precision training~\cite{micikevicius2017mixed} to reduce communication overhead. Another approach involves overlapping computation and communication~\cite{zhang2025comet} to further minimize training delays.

Storage systems are also critical to LLM training, particularly for efficiently saving and loading checkpoints. 
Tecionic~\cite{pan2021facebook} from Meta addresses frequent checkpoint writes. FireFlyer~\cite{an2024fire}, developed by DeepSeek, uses the custom-built 3FS distributed file system to manage checkpoint storage.

\textbf{Startup Research.}
Although startup performance has not been extensively studied in the context of LLM training, the literature in other domains offers several techniques that could help accelerate LLM training startup. 
For instance, prior work on Function-as-a-Service (FaaS) systems focuses on fast startup of relatively small containers and programs. 
Catalyzer~\cite{du2020catalyzer} improves sandbox restoration by efficiently resuming serverless functions across sandboxes, containers, and virtual machines. 
MITOSIS~\cite{wei2023no} extends this approach by introducing remote sandbox forking using Remote Direct Memory Access (RDMA). 
TrEnv~\cite{huang2024trenv} further applies CXL/RDMA-based remote memory pools to share entire execution environments across different functions. 
Additionally, infrastructure-level studies from AWS~\cite{brooker2023demand} and Microsoft~\cite{meyer2012study} explore techniques for accelerating container loading.

\section{Conclusion and Future Work}
\label{sec:conclusion}
In this work, we present the first comprehensive characterization of LLM training startup overhead using production data from over 28,000 jobs, revealing that startup processes can consume a substantial portion of GPU resources and hinder training efficiency. 
Based on our analysis, we design and implement \textbf{BootSeer}, a production-ready optimization framework that targets key startup bottlenecks: image loading, dependency setup, and checkpoint resumption, through caching, prefetching, and peer-to-peer sharing. 
BootSeer significantly reduces startup time by 50\%, eliminates straggler effects, and improves overall system responsiveness, demonstrating that targeted startup optimizations are essential for efficient and scalable LLM training.

\textbf{Co-designing Environment Caching with RDMA Networks.}  
RDMA is essential for LLM training but is typically unused during the startup phase.
(In principle, for example in inferencing, there could be another production job using RDMA.  But for training jobs, a single job usually occupies the entire physical machine.) By sharing the environment cache over RDMA, further startup optimizations can be achieved. For critical jobs, we can maintain the entire execution environment in a remote memory pool and share it across worker nodes using a peer-to-peer network and a copy-on-write mechanism.

\textbf{Process Snapshots to Accelerate Large Process Startups.}  
During startup, numerous daemon processes must be launched as prerequisites. By identifying these processes and creating snapshots of their initialized state, we can skip repetitive initialization, thereby reducing startup overhead.

\clearpage

\bibliographystyle{plainnat}
\bibliography{main}

\clearpage

\end{document}